\documentclass[%
 aip,
 amsmath,amssymb,
 reprint,%
]{revtex4-1}

\usepackage{graphicx}
\usepackage{dcolumn}
\usepackage{bm}

\usepackage[utf8]{inputenc}
\usepackage[T1]{fontenc}
\usepackage{mathptmx}
\usepackage{etoolbox}

\usepackage{listings}
\usepackage{adjustbox}
\usepackage{graphicx}
\usepackage{color}
\usepackage{caption}
\usepackage{comment}
\usepackage{amsmath}
\usepackage{amsfonts}
\usepackage{amssymb}
\usepackage{graphicx}
\usepackage{float}
\usepackage{placeins}
\usepackage{balance}
\usepackage{url}
\usepackage{algorithm2e}
\DeclareMathVersion{normal2}
\mathversion{normal2}
\usepackage{mathrsfs}
\mathversion{normal}
\usepackage{bm}

\makeatletter
\def\@email#1#2{%
 \endgroup
 \patchcmd{\titleblock@produce}
  {\frontmatter@RRAPformat}
  {\frontmatter@RRAPformat{\produce@RRAP{*#1\href{mailto:#2}{#2}}}\frontmatter@RRAPformat}
  {}{}
}%
\makeatother
\begin{document}

\preprint{AIP/123-QED}

\title[]{Quantifying Complexity: \\An Object-Relations Approach to Complex Systems}
\author{Stephen Casey}
\affiliation{ 
NASA Langley Research Center
}%
\email{Stephen.Casey@nasa.gov}

\date{\today}

\begin{abstract}
The best way to model, understand, and quantify the information contained in complex systems is an open question in physics, mathematics, and computer science.  The uncertain relationship between entropy and complexity further complicates this question.  With ideas drawn from the object-relations theory of psychology, this paper develops an object-relations model of complex systems which generalizes to systems of all types, including mathematical operations, machines, biological organisms, and social structures.  The resulting Complex Information Entropy (CIE) equation is a robust method to quantify complexity across various contexts.  The paper also describes algorithms to iteratively update and improve approximate solutions to the CIE equation, to recursively infer the composition of complex systems, and to discover the connections among objects across different lengthscales and timescales.  Applications are discussed in the fields of engineering design, atomic and molecular physics, chemistry, materials science, neuroscience, psychology, sociology, ecology, economics, and medicine.
\end{abstract}

\maketitle

\begin{quotation}
How can one observe the behavior of a system and determine its inner workings?  It is usually impossible to solve this problem exactly, since most real-world models contain uncertainty in relation to the systems they represent.  There is a tradeoff between model accuracy, model size, and computational cost.  There are often also hidden variables and elements of stochasticity.
\end{quotation}


\LinesNumbered
\RestyleAlgo{ruled}
\SetKwProg{Fn}{Function}{:}{end}
\SetKwInput{kwGet}{get}{}{}

\section{\label{sec:level1}Introduction}
Minimum Description Length (MDL) is often used as a parameter in model selection to fit a particular dataset, whereby the shortest description of the data, or the model with the best compression ratio, is assumed to be the best model \cite{rissanen1978modeling}.  MDL is a mathematical implementation of \textbf{Occam’s razor}, which dictates that among competing models, the model with the fewest assumptions is to be preferred.  If an MDL description can contain all of the operations in a Turing-complete programming language, the description length then becomes equivalent to the Kolmogorov complexity, which is the length of the shortest computer program that produces the dataset as output\cite{kolmogorov1965three}.  For a given dataset, a program is Pareto-optimal if there is no shorter program that produces a more accurate output.  The graph of Kolmogorov program length vs. accuracy is the Pareto frontier of a dataset.

If one combines Occam’s razor with the Epicurean \textbf{Principle of Multiple Explanations}, the result is Solomonoff's theory of inductive inference \cite{solomonoff1964formal1}\cite{solomonoff1964formal2}.   According to the Principle of Multiple Explanations, if more than one theory is consistent with the observations, all such theories should be kept.  Solomonoff's induction considers all computable theories that may have generated an observed dataset, while assigning a higher Bayesian posterior probability to shorter computable theories.  The theory of inductive inference uses a computational formalization of Bayesian statistics to consider multiple competing programs simultaneously in accordance with the Principle of Multiple Explanations, while prioritizing shorter programs in accordance with Occam's razor.

A characteristic feature of complex systems is their activity across a wide range of lengthscales and timescales.  As a consequence, they can often be divided into a hierarchy of sub-components.  In order to reduce the computational difficulty of programs that can be divided into sub-programs, the dynamic programming method, originally developed by Richard Bellman \cite{bellman1954theory}, simplifies a complicated problem by recursively breaking it into simpler sub-problems.   The \textbf{divide-and-conquer} technique used by dynamic programming can applied to both computer programming and mathematical optimization.  A problem is said to have optimal substructure if it can be solved optimally by recursively breaking into sub-problems and finding the optimal solution to each sub-problem.

In contrast to the divide-and-conquer procedure, a \textbf{unification} procedure finds a single underlying theory that can explain two or more separate higher-level theories.  Causal learning algorithms or structural learning algorithms are examples of unification procedures, which determine cause-and-effect among variables in an observed dataset.  These variables may be hidden or unknown, in which case they must be inferred via hidden Markov models, Bayesian inference models, or other methods.  

\begin{figure*}
    \centering
    \includegraphics[width=0.94\textwidth]{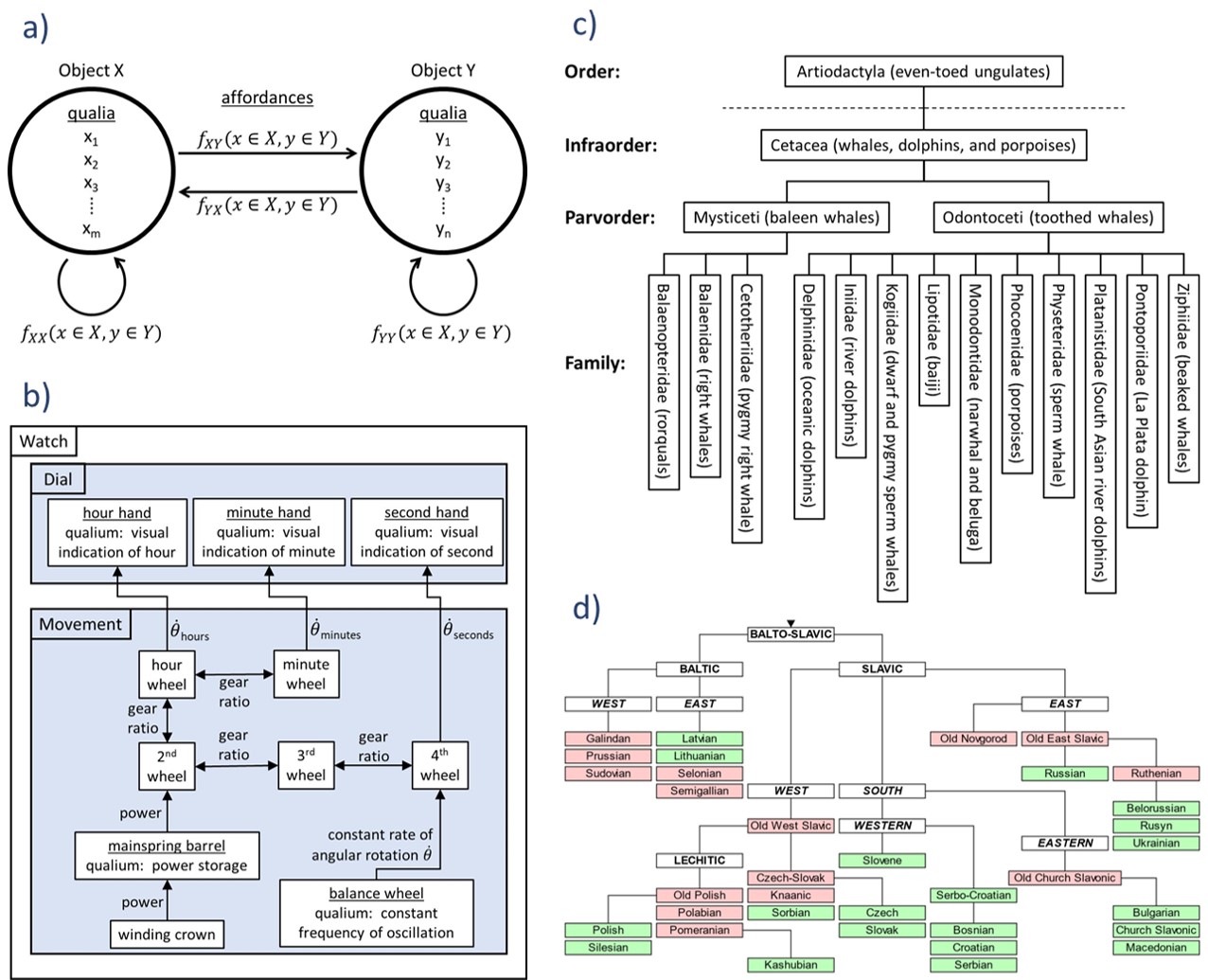}
    \caption{a) Objects contain qualia, which are properties, variables, or other objects.  Affordances are modes of interaction.  b) The relationships among objects can be shown using system diagrams.  In this example, a mechanical watch is composed of a dial object and a movement object.  The movement object is composed of gears, regulators, and springs, which affect each other through mechanical affordances.  c) The categories used for biological classification are not exact, but are hypothetical objects inferred by scientists and subject to revision.  This graph shows the categories for cetaceans between the ranks of order and family.   d) Connected objects in the graph share some amount of structure and mutual information.  In this language diagram, depicting the derivatives of the Balto-Slavic language which may have split into Baltic and Slavic around 1400 BCE\cite{gray2003language}, each language derives most of its information and structure from its predecessor in the graph.  Image licensed under CC BY-SA 3.0.}
    \label{fig:img58}
\end{figure*}

These four principles – Occam’s razor, the Principle of Multiple Explanations, divide-and-conquer, and unification – have been used in various combinations to analyze complex systems.  Some researchers have used them all together, such as Wu, Udrescu, and Tegmark in their endeavor to create an AI physicist\cite{wu2019toward}\cite{Udrescu_2019}\cite{udrescu2020ai2}.  In order to fully characterize the information contained in complex systems, a fifth item should be added to this list:  the principle of \textbf{overinterpretation}.  This principle states there can be multiple correct interpretations of the same dataset, and the validity of any particular lens of interpretation is determined by whether or not it produces useful insights.  This is distinct from the Principle of Multiple Explanations, which states that no plausible theory can be ruled out, but which still assumes there is one correct theory among many plausible.  In contrast to the Principle of Multiple Explanations, the principle of overinterpretation states that multiple interpretations can be simultaneously correct.  In the same way that the Principle of Multiple Explanations was suggested by Epicurus and refined by Bayes or Laplace, the principle of overinterpretation may be said to have been suggested by Freud in his analysis of Hamlet in The Interpretation of Dreams\footnote{“But just as all neurotic symptoms, and, for that matter, dreams, are capable of being ‘overinterpreted' and indeed need to be, if they are to be fully understood, so all genuinely creative writings are the product of more than a single motive and more than a single impulse in the poet's mind, and are open to more than a single interpretation.”\cite{freud1900interpretation}} and refined by Walter Kaufmann\cite{kaufmann1980}.  In a mathematical context, overinterpretation indicates there may be multiple non-equivalent Pareto-optimal objects within a given dataset.

\section{Theory}

The definitions used here are as general as possible.  The word \textbf{object} is defined as an entity containing two features:  1.  A set of properties, variables, or other objects, called \textbf{qualia} (the singular form used here is \textbf{qualium}), and 2.  functions between itself and other objects, called \textbf{affordances}.  A summary is shown in Figure \ref{fig:img58}a. 

An example for a mechanical watch is shown in Figure \ref{fig:img58}b.  The watch is composed of the dial and the movement, which are themselves composed of constituent objects.  Systems engineers and computer scientists will recognize the similarities to system design documents and Unified Modeling Language (UML) diagrams.

\subsection{Hierarchical Graphs}

In addition to representing objects using component diagrams or Venn diagrams, it is often useful to employ graph decompositions.  A taxonomy diagram for cetaceans is shown in Figure \ref{fig:img58}c, which separates general categories into more specific categories.  Similarly, Figure \ref{fig:img58}d shows the evolutionary diagram of the Balto-Slavic languages.  These graphs are hierarchical in the sense that each object derives much of its information and structure from its predecessor.

\subsection{Constructing and Deconstructing Objects}

What exactly constitutes an object?  Roughly, an object is an information-theoretic construct that contains a lot of information but does not require much description.  Shea\cite{Shea_2021} has applied this definition in physics by analyzing a fluid system containing both laminar and turbulent flow.  Without preprogrammed knowledge of different flow regimes, his machine learning algorithm discovered the boundary between the laminar and turbulent zones by minimizing the complexity of the model relative to its predictive capability.

\begin{figure}[ht!]
    \includegraphics[width=\linewidth]{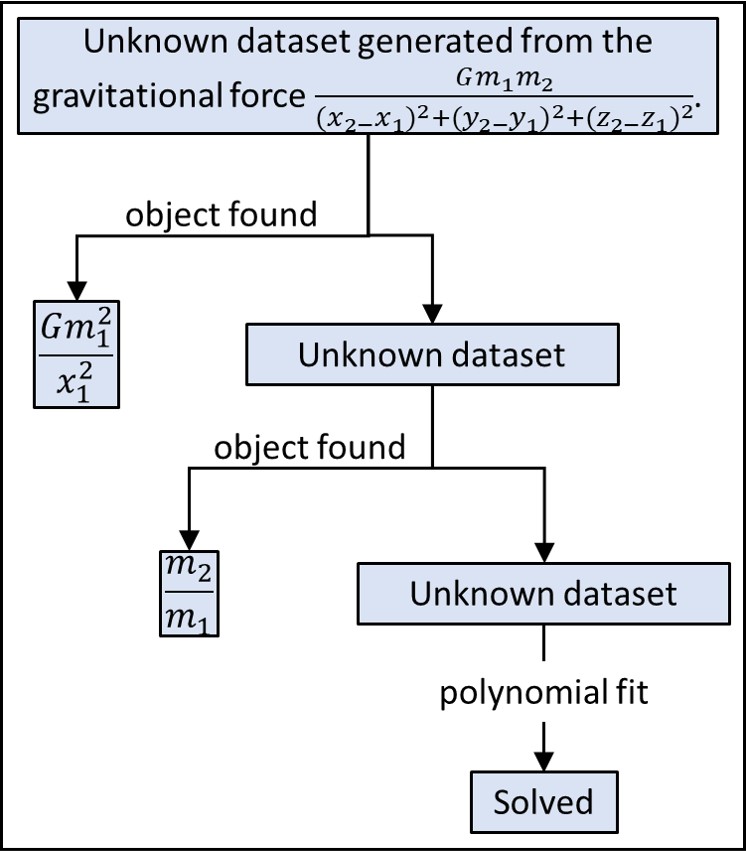}
    \caption{ML is used to procedurally find objects within a table of simulated gravitational measurements.  Image adapted from Figure 2 in \cite{Udrescu_2019}.  Used with permission.}
    \label{fig:img43}
\end{figure}

In simple algebraic systems, objects can be symbolic variables while affordances are mathematical functions.  Udrescu\cite{Udrescu_2019} has employed this formulation in the development of a machine learning algorithm to discover well-known physical laws from tables of experimental data.  Figure \ref{fig:img43} shows this process for Newton's law of universal gravitation $F = Gm_1m_2 / r^2$.  Given a table of $G,m,x,y$ and $z$ values, the program first discovers the dimensionless quantity $Gm_1^2/x_1^2$, then the ratio of masses $m_2/m_1$, then finally discovers the relationship among the position variables via polynomial regression.  The uncertainty present in the dataset is reduced each time a new object is found.

\subsubsection{Structural Learning Methods}

Much previous work in learning the structural composition of systems has focused on Bayesian belief networks and hidden Markov models.  A Bayesian belief network is a probabilistic model that represents a set of variables and their conditional dependencies via a directed acyclic graph, meaning a graph in which information flows in only one direction along each edge, and which do not contain any circular causation loops.  These graphs are ideal for predicting the probable causes and contributing factors of observed events.  Algorithms exist for inferring unobserved variables, in addition to algorithms for belief propagation which can conditional probabilities throughout the graph in response to new evidence.

A hidden Markov model (HMM) assumes the system under investigation is a Markov process, which is a network of possible states of the system and the probabilities of transitions between states.  HMM's attempt to infer unobserved hidden states of the system.  The graphs used for Markov random fields can be undirected and contain cycles, which are more general than the directed acyclic graphs used for belief nets.  Consequently, HMM's are a useful analysis tool that have found applications across numerous fields.

While a great deal of progress has been made in structural inference algorithms such as Bayesian belief networks and hidden Markov models, they are limited in their ability to infer relationships among variables beyond conditional probabilities.  Conway's Game of Life is a good example.  The Game of Life is a well-known cellular automaton devised by John Conway\cite{gardner1970fantastic} which uses simple rules to produce complex and lifelike behavior on a two-dimensional grid of square cells using discrete timesteps.  The game applies the following rules each timestep:

\begin{enumerate}
    \item Any live cell with fewer than two live neighbors dies.
    \item Any live cell with two or three live neighbors lives.
    \item Any live cell with more than three live neighbors dies.
    \item Any dead cell with exactly three live neighbors becomes a live cell.
\end{enumerate}

Although these rules are simple, they will defeat a probabilistic inference algorithm's attempt to discover them.  This is because the behavior of a cell is not stochastic in relationship to its neighbors, but instead is based on the application of basic logical operations.  In order to solve this system, something more akin to Wu, Udrescu, and Tegmark's AI physicist is needed.

\subsubsection{The Role of Machine Learning}
\label{section:ml}

\begin{figure}[t!]
    \includegraphics[width=\linewidth]{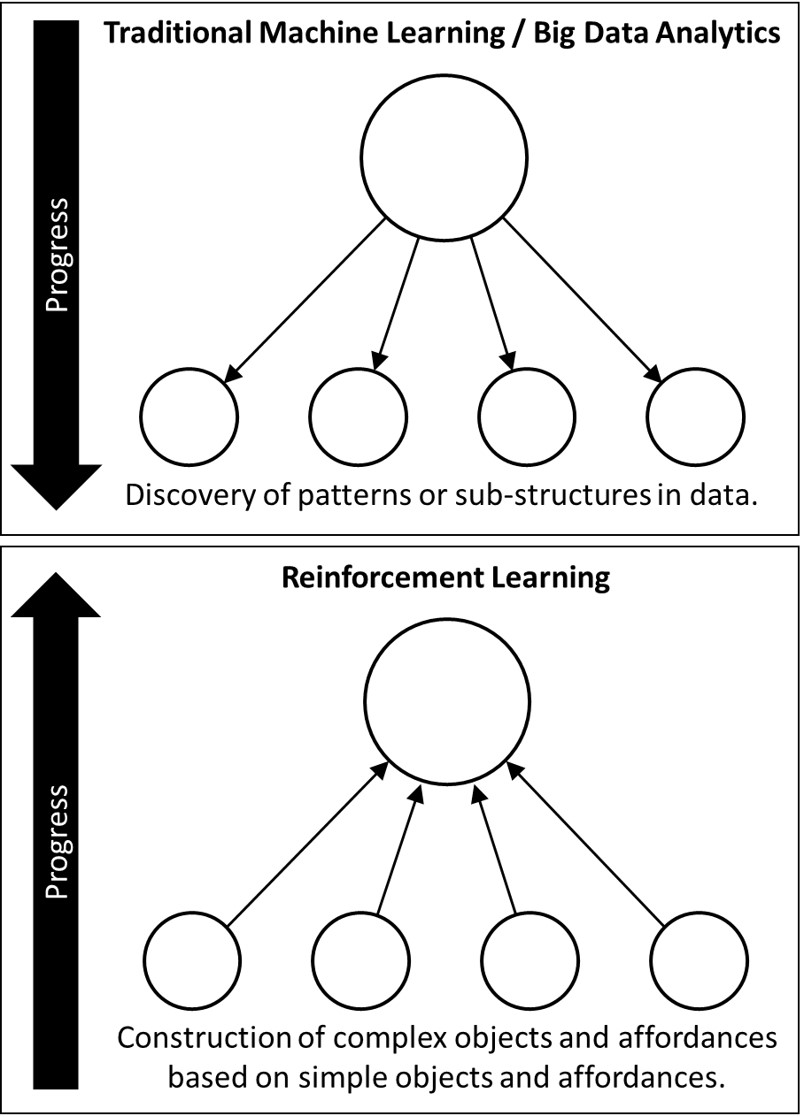}
        \caption{Machine learning involves either the deduction of objects present in datasets (top) or the construction of complex objects from simple constituents (bottom).}
    \label{fig:img44}
\end{figure}

Continuing with the Game of Life example, a more robust machine learning (ML) algorithm can be devised to infer objects and affordances at across a range of length- and timescales.  The various subtypes of ML can be broadly divided into two categories.  Casual references to machine learning or data science typically refer to the first category, termed here as traditional ML or big-data analytics.  This category involves analyzing large datasets using some combination of statistical methods to uncover previously unknown patterns or relationships among variables.  Image recognition would be included in this group, as the neural network must analyze pixels in an image to determine unknown variables, such as whether the image contains a dog or a cat.

The second subtype of machine learning is reinforcement learning (RL), wherein an agent interacts with its environment in order to achieve a goal or set of goals.  Much RL development has focused on games, such as chess, go, and classic Atari video games.  In these environments, the agents use a finite set of actions, such as chess moves or video game inputs, to progress toward achieving a long-term goal.  In this process, RL agents often discover advantageous techniques or game configurations already known to human experts, then proceed to surpass human experts by discovering previously unknown techniques.  In contrast to the traditional ML category, the RL category involves the construction of advanced objects and affordances from simpler objects and affordances.  This dichotomy is illustrated in Figure \ref{fig:img44}.

\begin{figure}[b!]
    \includegraphics[width=\linewidth]{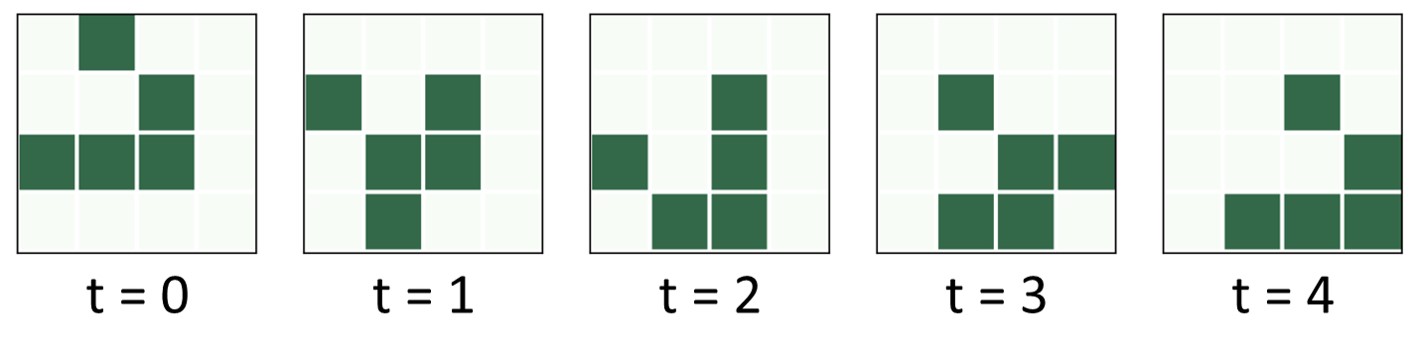}
    \caption{The glider in Conway's Game of Life is a pattern that moves diagonally down and right.  When viewed from the macroscale, the glider is an object moving across the game board.  However, when viewed on the microscale, it's a collection of tiles turning on and off.  Image licensed under Creative Commons\cite{glider}.}
    \label{fig:img55}
\end{figure}

The glider in Figure \ref{fig:img55} is one of the of the simple forms found in the Game of Life.  As illustrated in the figure, the glider moves one space down and to the right every five timesteps.  The glider appears to be moving across the game board when viewed from a distance.  However, when viewed at the level of individual cells, motion is not possible as the cells can only turn on and off.  The cells do not have the affordance of motion, but this affordance appears through the collective action of multiple objects.  This emergence of complex affordances from combinations of simple affordances is the process denoted by the bottom half of Figure \ref{fig:img44}.  As with many physical systems, there is a difference in the behavior and understanding of the glider at the microscale compared to the macroscale.

Direct parallels can be drawn between the simple Game of Life example and problems in physics and engineering.  For example, the prediction of macroscopic chemical properties based on molecular structure is challenging due to the complex behavior caused by quantum effects on the microscale.  A similar problem exists in neuroscience in the attempt to bridge the gaps between the behavior of individual neurons and the brain en masse.

\begin{algorithm}[t]
\caption{Zoom-in-Zoom-out (ZIZO)}\label{alg:zizo}
    \SetKwFunction{Fzo}{Zoom\_out}
    \SetKwProg{Fn}{Function}{:}{\KwRet $X$}
    \Fn{\Fzo{Y}}{
        \tcc{This function takes a top-down approach from the macroscale.}
        View the system as a black box, and infer the hidden objects X that could be causing its behavior\;
        If Y is provided, condition this process on Y\;
    }
    \SetKwFunction{FDeduce}{Zoom\_in}
    \SetKwProg{Fn}{Function}{:}{\KwRet $Y$}
    \Fn{\FDeduce{X}}{
        \tcc{This function takes a bottom-up approach from the microscale.}
        Encode the microstructure of the subsystem into representative objects Y\;
        If X is provided, condition this process on X\;
    }
	\SetKwFunction{FMain}{}
	\SetKwProg{Pn}{Main}{:}{\KwRet}
	\Pn{\FMain{}}{
	    \While{$X \neq Y$}{
            X = Zoom\_out(Y)\;
            Y = Zoom\_in(X)\;
    	    Loop until convergence\;
        }
    }
\end{algorithm}

In order to address this problem, we have defined an algorithm to discover the connections among objects across different lengthscales and timescales.  Inspired by the Grow-Shrink algorithm used for Bayesian network structure learning and the forward-backward algorithm used for Markov models, here we have defined a Zoom-in-Zoom-out (ZIZO) algorithm for object-relations models.  The Grow-Shrink algorithm works by adding variables to nodes in a Bayesian network graph until all dependencies have been captured (the ``grow" phase), then tests these dependencies and removes superfluous connections (the ``shrink" phase)\cite{margaritis2003learning}.  The forward-backward algorithm for HMM's calculates the probability distribution of the system's hidden state variables for a sequence of observations moving forward in time, then repeats the process for the sequence in reverse, with the intention to cause the forward-in-time model and the backward-in-time model to converge\cite{rabiner1986introduction}.  The ZIZO algorithm performs a similar type of oscillation by first observing a phenomenon from a zoomed-out, macroscopic viewpoint and developing a top-down model, then zooming into an object on the microscopic level and building a bottom-up model, then iterating between the two models until they converge on a common set of objects shared between the two.

Note the ZIZO algorithm has so far only been discussed as a method to describe complex systems.  However, if the algorithm is used in a reinforcement learning mode, and the program tries to create desired affordances rather than trying to model observations, then the algorithm changes from a tool of system description to a tool of system design.  This type of tool could be used to autonomously design and optimize all types of engineering systems, perhaps beyond the capabilities of human experts.  The introduction of intelligent sampling via active learning, in addition to simulation capabilities, could further improve the utility of this process.

\subsection{A Simple Example:  A Connected Graph with 8 Nodes}

\begin{figure*}
    \centering
    \includegraphics[width=1\textwidth]{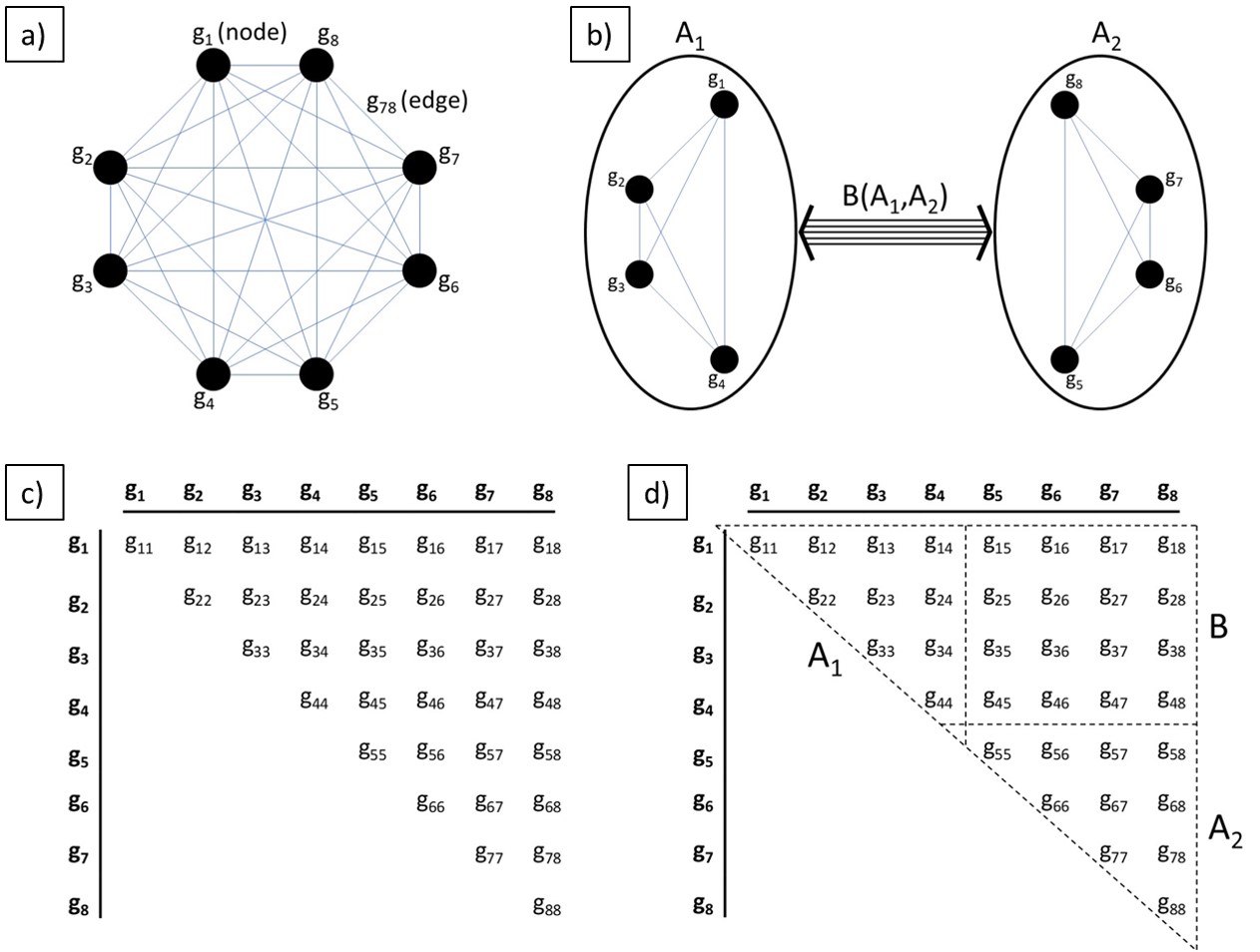}
    \caption{a) Graph $G$ is composed of nodes $g_1$ through $g_8$ and possible edges $g_{ij}$ between nodes $g_i$ and $g_j$ for $i,j \in [1,8]$.  b) Dividing $G$ into two objects $A_1$ and $A_2$ produces two k-connected graphs of size 4 and a function $B$ that communicates between them.  c) As $g_{ij}=g_{ji}$ the adjacency matrix of $G$ is symmetric about the diagonal.  d) Dividing $G$ into $A_1$ and $A_2$ has the effect of segmenting the adjacency matrix.}
    \label{fig:img62}
\end{figure*}

As an intuitive introduction to the theory of delineating objects, we will define a graph $G$ as a k-connected graph of size 8.  There are 8 nodes, each of which may or may not share an edge with any of its neighbors or with itself.  We will assume edges are bidirectional, meaning edge $g_{12}$ is the same as edge $g_{21}$.  Figure \ref{fig:img62}a illustrates this graph.

There may or may not be an edge $g_{ij}$ between nodes $g_{i}$ and $g_{j}$.  If there is an edge, this condition is represented by $g_{ij}=1$.  If not, $g_{ij}=0$.  We will assume no prior knowledge about the existence of each edge, meaning $g_{ij}$ may be 0 or 1 with equal probability.  The full information contained in $G$ is therefore represented by the adjacency matrix shown in Figure \ref{fig:img62}c.

The adjacency matrix contains 36 independent binary variables.  This translates to $2^{36}$ equally probable configurations of $G$.  The Shannon entropy of $G$ is therefore 
\begin{equation}
    H(G) = \mathrm{log}_2 2^{36} = 36 \mathrm{\ bits}.
\end{equation}
Intuitively, information entropy can be considered the average amount of information conveyed by an event when considering all possible outcomes.

Let us now consider what happens if $G$ is not regarded as a single object, but is instead analyzed as two objects $A_1$ and $A_2$ as shown in Figure \ref{fig:img62}c.  $A_1$ is constructed to contain nodes 1-4, while $A_2$ contains nodes 5-8.  An affordance $B$ describes the possible modes of interaction between $A_1$ and $A_2$.

Using the boundaries drawn in this example, $A_1$ and $A_2$ are k-connected graphs of size 4.  Since $A_1$ contains nodes 1-4, the self-entropy of $A_1$ is the entropy contained in the edges $g_{ij}$ for $i,j \in [1,4]$.  Likewise, the self-entropy of $A_2$ is the entropy contained in the edges $g_{ij}$ for $i,j \in [5,8]$.  The function $B$ must contain information about the connections between nodes 1-4 and nodes 5-8, therefore $B$ contains the values of $g_{ij}$ for $i \in [1,4]$ and $j \in [5,8]$.
\\

These groups are visualized on the adjacency matrix in Figure \ref{fig:img62}b.  The separation of $G$ into $A_1$ and $A_2$ had the effect of dividing the adjacency matrix into sections.  The information content of each section can be analyzed separately:
\begin{align}
    H(A_1) = \mathrm{log}_22^{10} = 10\\
    H(A_2) = \mathrm{log}_22^{10} = 10\\
    H(B) = \mathrm{log}_22^{16} = 16.
\end{align}
Total information in the system is conserved:
\begin{equation}
    \label{eqn:eqn1}
    H_{\mathrm{total}} = H(G) = H(A_1)+H(A_2)+H(B) = 36.
\end{equation}
Conditional entropy equalities also hold.  For example,
\begin{equation}
    H(A_1) = H(G) - H(G \mid A_2,B).
\end{equation}

\subsubsection{Discussion}
In this example, it's clear dividing $G$ into two connected objects is equivalent to considering $G$ as a single object.  This can be verified by the equality in Equation \ref{eqn:eqn1} or by visually inspecting Figure \ref{fig:img62}d.  In fact, the boundaries around $A_1$ and $A_2$ can be drawn around any set of nodes and the outcome will be equivalent in regard to the total system entropy.  The only change will be how the adjacency matrix is divided into subsections.

From an intuitive standpoint, the separation of $G$ into $A_1$ and $A_2$ seems undesirable, as a large amount of information needs to be contained in the connection $B$.  $B$ contains 16 bits of information, while the objects $A_1$ and $A_2$ each contain only 10.  Separating $A_1$ and $A_2$ does not decrease the amount of information needed to describe the system.

However, what if the values in $B$ are almost always zero?  In this case, the entropy contained in $B$ is very low.  If nodes 1-4 are almost never connected to nodes 5-8, it becomes advantageous to consider graphs $A_1$ and $A_2$ as separate groups.

\subsection{Complex Information Entropy}

\begin{figure*}
    \centering
    \includegraphics[width=1\textwidth]{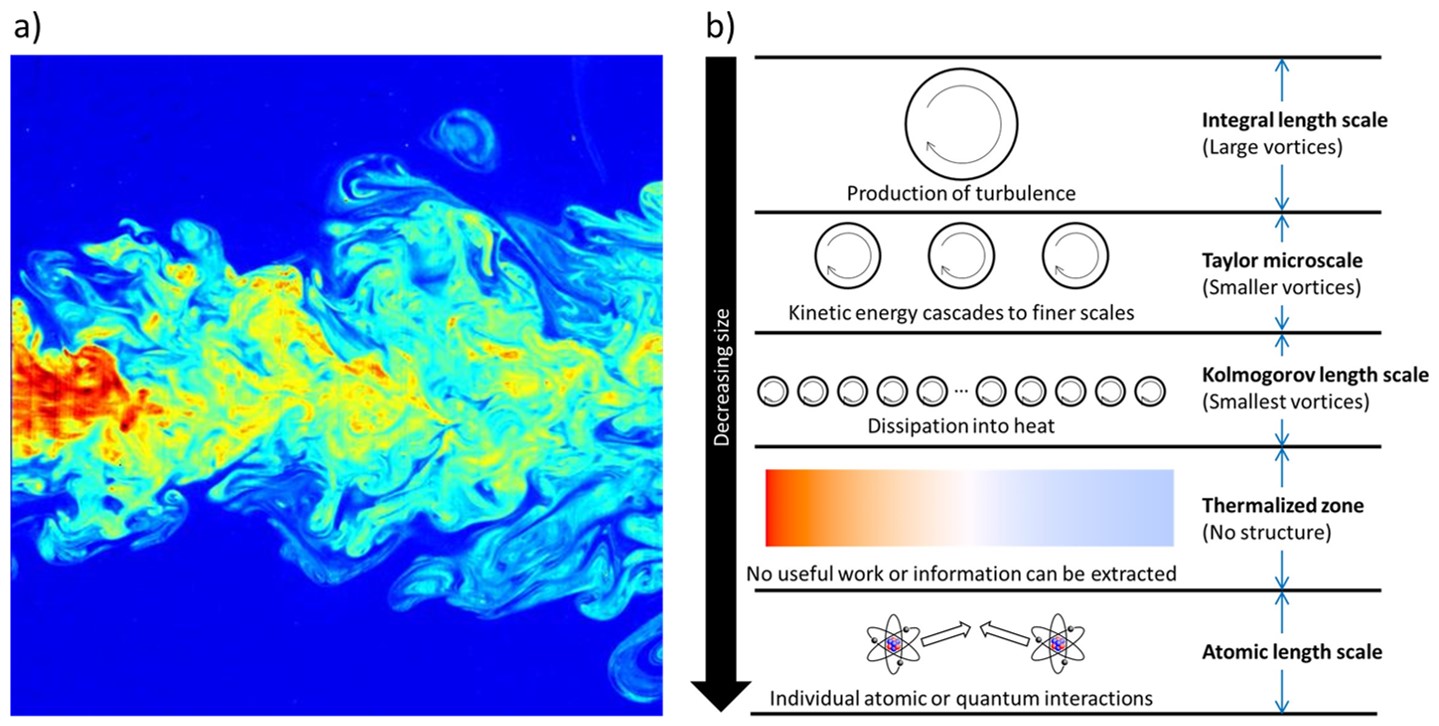}
    \caption{a) False-color flow visualization of a submerged turbulent jet, made by laser-induced fluorescence.  Imaged licensed under CC BY-SA 3.0\cite{turb}.  b) The CIE of turbulence is high due to presence of structures across man lengthscales and timescales.}
    \label{fig:img60}
\end{figure*}

Here we will define a useful method to measure the information entropy of any system.  The Complex Information Entropy (CIE) of a system is defined as

\begin{equation}
    \label{eqn:cie}
    \begin{gathered}
        \textrm{Complex Information Entropy } \mathnormal{}C^*(S) = \\
        \sum_{A_i \in \mathbb{A^*}}c_iH(A_i) + \underset{B_{ij} \in \mathbb{B}^*\mathnormal{}}{\sum_{A_i, A_j \in \mathbb{A^*}, \ \mathnormal{}}}d_{ij}H(B_{ij}(A_i, A_j))
    \end{gathered}
\end{equation}

\noindent where $\mathbb{A}^*$ and $\mathbb{B}^*$ are the sets of all Pareto-optimal objects and affordances in system $S$, $A_i$ and $B_{ij}$ are defined in Equation \ref{eqn:po}, and $c_i$ and $d_{ij}$ are empirically-determined coefficients used to adjust for information not captured by $A_i$ or $B_{ij}$.  This function is recursive, as each object $A_i$ may itself be considered a complex system, and $H(A_i)$ can be calculated as the CIE of the objects and affordances inside $A_i$.  Thus the CIE function accounts for the hierarchical representation of information in complex systems.

$C^*$ indicates the optimum value of $C$ given a complete set of optimized objects and affordances.  Similar to the uncomputability of Kolmogorov complexity, meaning there is no finite-length program that will return the Kolmogorov complexity for a given input string, $C^*$ may also be uncomputable.  In practice, real-world programs may be used to iteratively find better values for $C$ over time.

Pareto-optimal objects and affordances are those that optimally simplify some subset $S'$ of $S$, as given by the equation

\begin{equation}
    \label{eqn:po}
    \begin{gathered}
        \mathnormal{}C(S' \subseteq S) = \\
        \underset{A_i \in \mathbb{A}, \ \mathnormal{}B_{ij} \in \mathbb{B}}{\mathrm{argmin}} \sum_{i}c_iH(A_i) + \sum_{i,j}d_{ij}H(B_{ij}(A_i, A_j))
    \end{gathered}
\end{equation}

\noindent where $\mathbb{A}$ and $\mathbb{B}$ are the sets of all possible objects and affordances in system $S$, including those that are sub-optimal.

\subsection{A More Complex Example:  Turbulence}

Equation \ref{eqn:cie} will now be applied to turbulent flow.  The best way to model, understand, and quantify the information contained in turbulence is an open question in physics.  The CIE metric provides useful insights when applied to such systems.

Turbulence possesses the property of structure at multiple scales.  Instabilities in the fluid flow form vorticies, which create secondary instabilities and smaller vorticies, which create even smaller vorticies, in a process that cascades to smaller length scales until the kinetic energy dissipates into heat.  The formation of these eddies can be seen in Figure \ref{fig:img60}a.

Researchers investigating the behavior of turbulent flow have identified three distinct physical regimes:  the integral length scale, the Taylor microscale, and the Kolmogorov length scale, as shown in Figure \ref{fig:img60}b.  The integral length scale is the regime in which eddies form and gather energy.  Below the integral length scale is the Taylor microscale in which energy is transferred from larger vortices to smaller vortices without being lost to the surrounding fluid.  Smallest is the Kolmogorov lengthscale, in which the viscosity of the fluid causes the energy of the vortices to be dissipated as heat.  There is effectively no structure below this scale, as the fluid has thermalized into a maximum-entropy condition characterized by the Maxwell-Boltzmann distribution from which no information can be extracted.

In addition to the energy differences, the three regimes also differ in their isotropic characteristics.  The integral length scale is highly anisotropic, meaning the eddies favor certain orientations in 3-dimensional space.  The Kolmogorov scale is isotropic, meaning the eddies occur evenly in every direction.  The transition from anisotropy to isotropy occurs in the Taylor microscale.

A lot of information is needed to fully describe a turbulent system due to the differences in each regime.  The CIE value for a system using the integral-Taylor-Kolmogorov schema is

\begin{equation}
    \label{eqn:eqn2}
    \begin{gathered}
        C(\mathrm{turbulence}) = \\ 
        C(\mathrm{Kolmogorov}) + C(\mathrm{Taylor}) + C(\mathrm{integral }) = \\
        \sum_{A_i \in \mathrm{Kol.}}c_iH(A_i) + \sum_{A_i, A_j \in \mathrm{Kol.}}d_{ij}H(B_{ij}(A_i, A_j)) + \\
        \sum_{A_i \in \mathrm{Kol.}, A_j \in \mathrm{Tay.}}d_{ij}H(B_{ij}(A_i, A_j)) + \\
        \sum_{A_i \in \mathrm{Tay.}}c_iH(A_i) + \sum_{A_i, A_j \in \mathrm{Tay.}}d_{ij}H(B_{ij}(A_i, A_j)) + \\
        \sum_{A_i \in \mathrm{Tay.}, A_j \in \mathrm{int.}}d_{ij}H(B_{ij}(A_i, A_j)) + \\
        \sum_{A_i \in \mathrm{int.}}c_iH(A_i) + \sum_{A_i, A_j \in \mathrm{int.}}d_{ij}H(B_{ij}(A_i, A_j)).
    \end{gathered}
\end{equation}

Equation \ref{eqn:eqn2} may appear to be unwieldy,  however its modularity is a major strength.  The CIE of each layer may be calculated separately.  Regardless of the number of layers in the hierarchy, each layer only needs to consider the interactions between itself and the layers immediately above and below.  Moreover, each object only needs to consider itself and the interactions with its immediate neighbors.

\begin{figure}[t]
    \includegraphics[width=\linewidth]{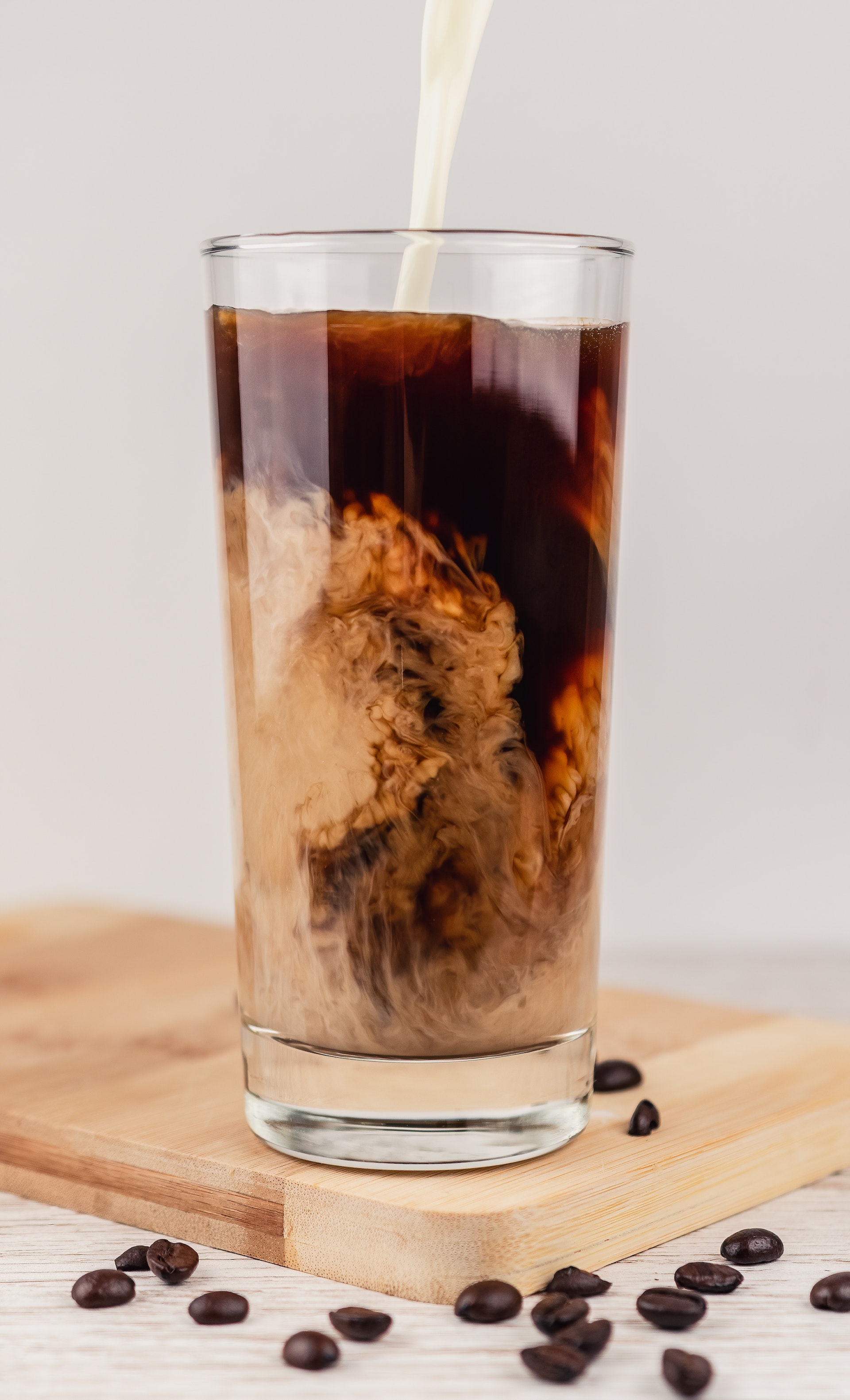}
    \caption{Mixing cream into coffee creates turbulent eddies at multiple lengthscales.  The complexity of the system is greatest when the two are partially mixed, although the classical Gibbs entropy is maximized when the mixture is homogeneous and without structure.  Imaged licensed under CC\cite{coffee}.}
    \label{fig:img41}
\end{figure}

As a sanity check, CIE behaves as desired for the classical example of cream being mixed into black coffee.  The Gibbs entropy used in statistical mechanics is maximized when the liquids are fully mixed and contain no internal structures, while the complexity is maximized during the mixing process when turbulent structures are present as shown in Figure \ref{fig:img41}.  As expected, CIE is low during the unmixed and fully-mixed states, and high during the partially-mixed state.

\subsection{The Virus Effect:  Propagation of Entropy through a System}

One feature of complex systems is the ability for small-scale changes to propagate throughout the system and cause large-scale effects.  For example, a person's exposure to a single virus, which is much smaller than a cell, can lead to a cascade of effects that overwhelm their entire body.  How can this be analyzed in an object-relations model?

The answer is found in the state entropy of each object.  Using the example of a virus, one may imagine the human body as consisting of a hierarchical arrangement of organs, tissues, and cells.  The physical processes and relevant length- and timescales for small components, such as cells, are much different than those for large components, such as organs.  Accordingly, the operation of organs is not affected by individual cells, and it is not necessary to model individual cells in order to accurately model organs.

If the output and states of object $A$ are discrete, the entropy of $A$ is given by the equation

\begin{equation}
    \label{discrete_entropy}
    H(A) = -\sum_{a \in \mathscr{A}} p(a)\mathrm{ln}\mathnormal{}(p(a)) = E[-\mathrm{ln}\mathnormal{}(p(A))]
\end{equation}

\noindent where $a$ is the internal condition of $A$, $\mathscr{A}$ is the set of all possible values of $a$, and $p(a)$ is the probability of occurrence of $a$.  If the states of $A$ are continuous, the entropy is given by

\begin{equation}
    \label{continuous_entropy}
    H(A) = E[-\mathrm{ln}\mathnormal{}(f(A))] = -\int_{\mathscr{A}} f(a)\mathrm{ln}\mathnormal{}(f(a))da.
\end{equation}

\noindent Here $\mathscr{A}$ is the support of $A$, meaning the domain over which $a$ is nonzero, and $f(A)$ is the probability density function of $A$.

If the scheme for representing the information in object $A$ is efficient, then the states that occur most frequently will carry the lowest information entropy, while the most unusual states will carry the highest information entropy.  In general, if $p_s$ is the probability of object $A$ being in state $s$, the entropy of object $A$ in state $s$ is $H(A(s)) = -p_s \mathrm{ln} (p_s)$.  In a living system such as a cell, the states expected to occur most frequently are ideally the states in which the cell is functioning as usual without any major problems.  If the cell is attacked by a deadly virus, the cell will be placed in a highly unusual state, which will carry a high amount of information $H(A)$.  If cell $A_i$ infects cell $A_j$ via $B_{ij}(A_i,A_j)$, then the information entropy of $A_j$ will also increase.

In this optimal multi-scale model of the body, if a hypothetical infection remains local, then the high-CIE state will also remain limited.  However, if the high-CIE state propagates to larger lengthscales, then tissues, organs, and eventually the entire body will be affected.  If the full body is in an abnormally high-CIE state, this could signal a medical emergency.

This type of evaluation could hypothetically be used to identify one harmful virus out of millions of harmless viruses.  Notably, such a model remains computationally tractable, as each object only needs to consider the interactions with the objects to which it is directly connected.

\subsection{Implications for Biological Organisms}

\begin{figure}[b]
    \includegraphics[width=\linewidth]{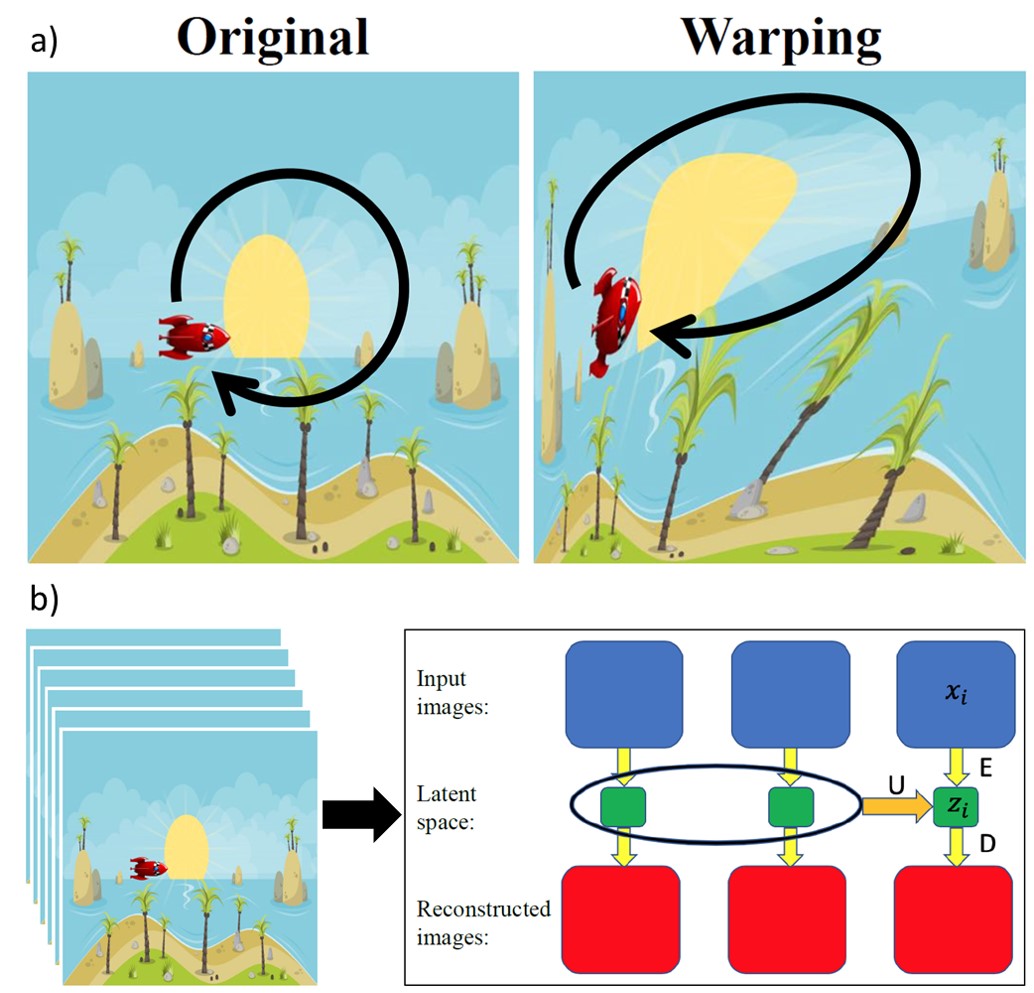}
    \caption{Using an autoencoder on distorted video resulted in the recovery of the original equations of motion for the rocket.  It is conjectured that humans perform a similar type of autoencoding of our perceptions in order to simplify our model of the world while retaining important information.  Image adapted from Figures 1, 2, and 3 in \cite{Udrescu_2021}.  Used with permission.}
    \label{fig:img50}
\end{figure}

\begin{algorithm*}[hbt!]
\caption{Complex System Structure-Learning Algorithm}\label{alg:structure_learning}
    \SetKwFunction{FMain}{}
	\SetKwProg{Pn}{Main}{:}{\KwRet}
	\Pn{\FMain{}}{
        \While{not converged}{
    	    Search for objects and affordances in the system\tcc*[r]{\textbf{divide-and-conquer}}
    	    Verify found objects and affordances are Pareto-optimal\tcc*[r]{\textbf{Occam's razor}}
    	    Add all Pareto-optimal objects and affordances found to a list:\tcc*[r]{\textbf{overinterpretation}}
    	    Check if objects in the list can be merged by using a common substructure\tcc*[r]{\textbf{unification}}
    	    Consider each object on the list to be its own system and repeat the above steps\tcc*[r]{\textbf{recursion}}
    	    Graph building:  search for dependencies among objects\;
    	    \If{If new dependencies are found}{
                Modify the objects and affordances if needed\;
    	    }
    	    Graph pruning:  check to see if existing dependencies can be removed:\;
    	    Loop until convergence, or repeat indefinitely\;
    	}
	}

\end{algorithm*}

The nervous systems of biological organisms seem to perform something conceptually similar to Equations \ref{eqn:cie} and \ref{eqn:po}, by creating a model of the world that is maximally descriptive while simultaneously being as simple as possible.  This is the Good Regulator theorem of cybernetics, which states that every good regulator of a system - meaning a device that maintains itself in a desired state - must contain a model of that system\cite{conant1970every}.  Sensory inputs are processed by the nervous system in order to construct an internal model of the world and to inform the organism's actions.  This is equivalent to the $\underset{A_i \in \mathbb{A}, \ \mathnormal{}B_{ij} \in \mathbb{B}}{\mathrm{argmin}}$ function of Equation \ref{eqn:po}, which finds the ideal choice of objects to represent a particular system.  The survival of an organism depends on its ability to recognize important information, disregard unimportant information, and update its model of the world in response to new observations.

Udrescu \cite{Udrescu_2021} has performed a simulation related to this idea.  He started by developing a video of a flying rocketship with a beach and palm trees in the background as shown in Figure \ref{fig:img50}.  The rocket moves around the frame by following trajectories determined by the equations for physical forces such as gravity, electromagnetism, and harmonic oscillation.  The video was then warped and distorted to obscure the rocket's motion.  Udrescu was able to recover the rocket's equations of motion by feeding the video frames into an autoencoder E, applying an operator U to the autoencoder's latent space, and training the operator and autoencoder simultaneously to achieve the simplest possible equation for U.  As a result, the autoencoder effectively learned how to unwarp the image, while the operator learned the symbolic equations of motion in a standard Cartesian coordinate system.

The implication of this experiment is that the human nervous system performs something similar to the function of the CIE equation:  developing a model of the world that is maximally descriptive while simultaneously being maximally efficient.  The following section will implement a practical example of such a model using trajectory data.


\section{Methods}

\begin{figure*}
    \centering
    \includegraphics[width=.9\textwidth]{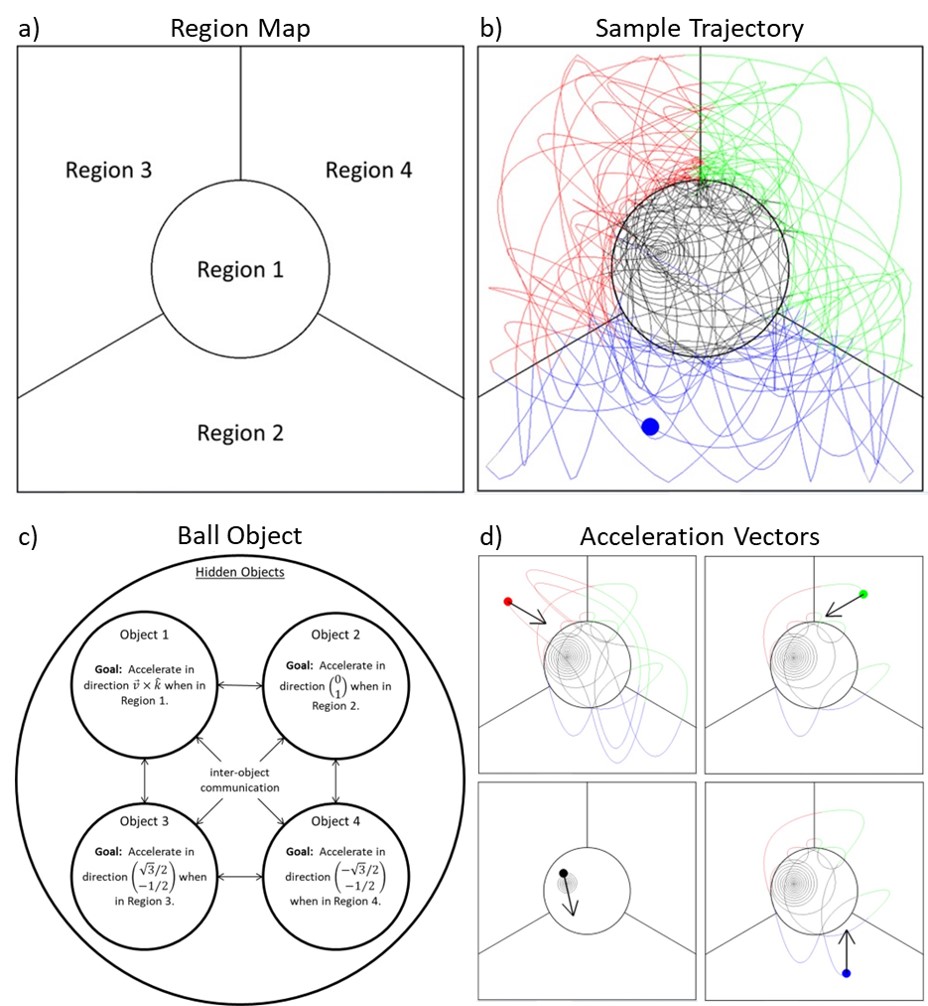}
    \caption{a) The map is divided into four regions.  The ball behaves differently in each region depending on which object is active.  b) The trajectory path is assigned a different color in each map region.  c) Four hidden objects work simultaneously to determine the ball's movement.  d) Acceleration vectors in each of the four regions.  The vector directions are constant in the edge regions and variable in the center.}
    \label{fig:img61}
\end{figure*}

The theory described above will now be used to discover hidden laws of motion within a set of trajectory data.  A general method for learning the structure of complex systems is shown in Algorithm \ref{alg:structure_learning}. This problem may be formulated as a reinforcement learning task and/or an optimization task, where the goal is to find the Pareto-optimal object boundaries.  Starting with some object $A$ with boundary $a = \partial A$, the goal is to iteratively adjust $a$ in order to Pareto-optimize all objects $A$ and affordances $B$.  The Bellman value equation for this problem takes the form

\begin{equation}
       C(s) = \underset{a = \partial A}{\mathrm{min}} \underset{s = \{A,B\}}{\sum} R(s,a,s') + \gamma C(s')
\end{equation}

\noindent where the state $s$ is the current state of the system model, $s'$ is the state from the previous iteration, $R$ is the reward for changing the model from state $s'$ to $s$ by changing the boundary $a$, $C(s')$ is the Complex Information Entropy from the previous state, and $\gamma$ is a factor which accelerates convergence\cite{bellman1957}.  Because the function is minimizing rather than maximizing $C$, $\gamma$ is greater than 1.  The corresponding update rule is
\begin{equation}
    C_{i+1}(s) \leftarrow \underset{a = \partial A}{\mathrm{min}} \underset{s = \{A,B\}}{\sum} R(s,a,s') + \gamma C_i(s').
\end{equation}
\noindent Ideally, this will produce an optimal policy for adjusting $a$.  In the following example, an annealing procedure is used as the method to adjust $A$ into an optimal configuration.

\subsection{Inferring Hidden Objects through Observation}

\begin{algorithm}[hbt!]
\caption{Deduction of Sub-Objects}\label{alg:deduction}
    \tcc{This program will try to continue finding objects in the dataset until its model of the data stops improving.}

    \SetKwFunction{FDeduce}{separate\_objects}
    \Fn{\FDeduce{dataset}}{
        Train a neural network on the complete dataset as a baseline\;
    	Randomly split the dataset into two subsets\;
    	Train two new NN's on each subset\;
    	\While{NN losses are decreasing}{
    	    Test each data point using both new NN's\;
    	    Reassign each data point to the set of the NN that fits best\;
    	    Retrain both NN's\;
    	    Loop until convergence\;
    	    \tcc{This is an annealing process.  Data points will migrate to subsets that minimize the combined losses from NN1 and NN2.}
    	}
    	\eIf{new NN losses $<$ baseline NN loss}{
    	    \KwRet the new subsets, new NN's, and total NN losses\;
    	    \tcc{New objects were found.}
    	}{\KwRet the original dataset, baseline NN, and baseline NN loss\;
    	    \tcc{New objects not found.}
    	    }
	}

	\SetKwFunction{FMain}{}
	\SetKwProg{Pn}{Main}{:}{\KwRet}
	\Pn{\FMain{}}{
	    \textbf{get} dataset\;
        \While{total NN losses are decreasing}{
            \ForEach{subset}{
    	        \textbf{call} separate\_objects(\textit{subset})\;
    	        \If{new objects are found}{
    	            Update the subsets for each object\;
    	        }
	        }
    	    Loop until convergence\;
        }
    }
\end{algorithm}

We begin with a square map divided into four regions as shown in Figure \ref{fig:img61}a.  A ball is placed at a random location on the map with a velocity vector pointed in a random direction with magnitude close to zero.  The ball then moves in a particular trajectory, depending on its region.  If the ball is in region 1, it accelerates in direction $\vec{v} \times \hat{k}$.  The direction of the acceleration vector in Region 2 is $\begin{pmatrix} 0\\ 1 \end{pmatrix}$, in Region 3 is $\begin{pmatrix} \sqrt{3}/2 \\ -1/2 \end{pmatrix}$, and in region 4 is $\begin{pmatrix} -\sqrt{3}/2 \\ -1/2 \end{pmatrix}$.  The magnitude of the acceleration is constant.  A sample trajectory is shown in Figure \ref{fig:img61}b.

In this simulation, external forces do not act on the ball.  Instead, the ball has a constant internal supply of energy and moves of its own volition depending on the region in which it finds itself.  The ball contains four hidden objects, each with a set of movement instructions, as shown in Figure \ref{fig:img61}c.  Communication between the hidden objects is used to allocate the ball's fixed supply of energy.

After running the simulation to generate the ball's trajectory, the $x$ and $y$ coordinates of the ball at each timestep can be used as a dataset to train a machine learning algorithm. Algorithm \ref{alg:deduction} shows the method used to infer the presence of hidden objects.  Each time a hypothetical new object is added, the performs an annealing process refine the model of each object.  The program continues to add objects until the addition of a new object does not improve the model's accuracy.

\section{Results}

\begin{figure}[b]
    \includegraphics[width=\linewidth]{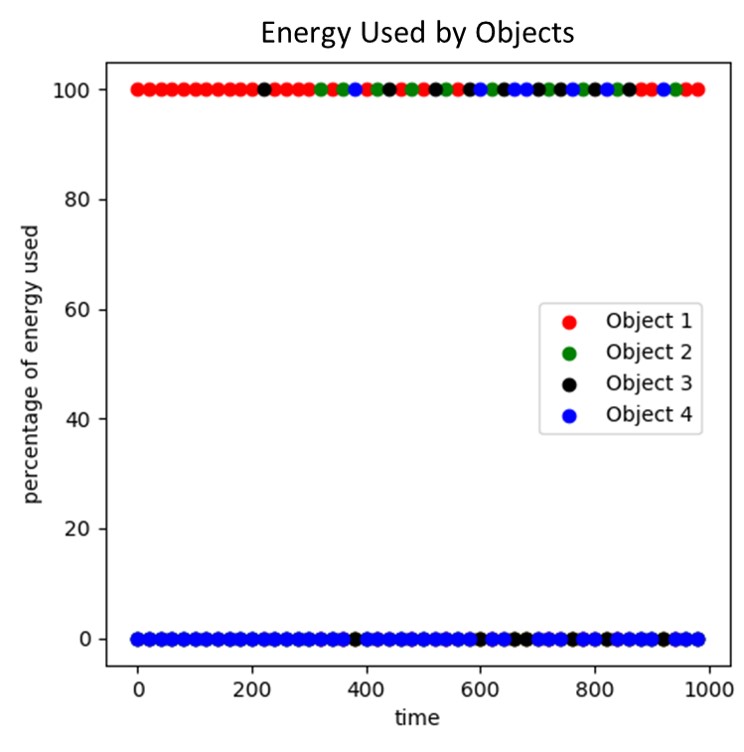}
    \caption{The object activations can be used to verify and gain insight into the model.  Only one object is active at a time during the course of the simulation.}
    \label{fig:img56}
\end{figure}

\begin{figure*}
    \centering
    \includegraphics[width=1\textwidth]{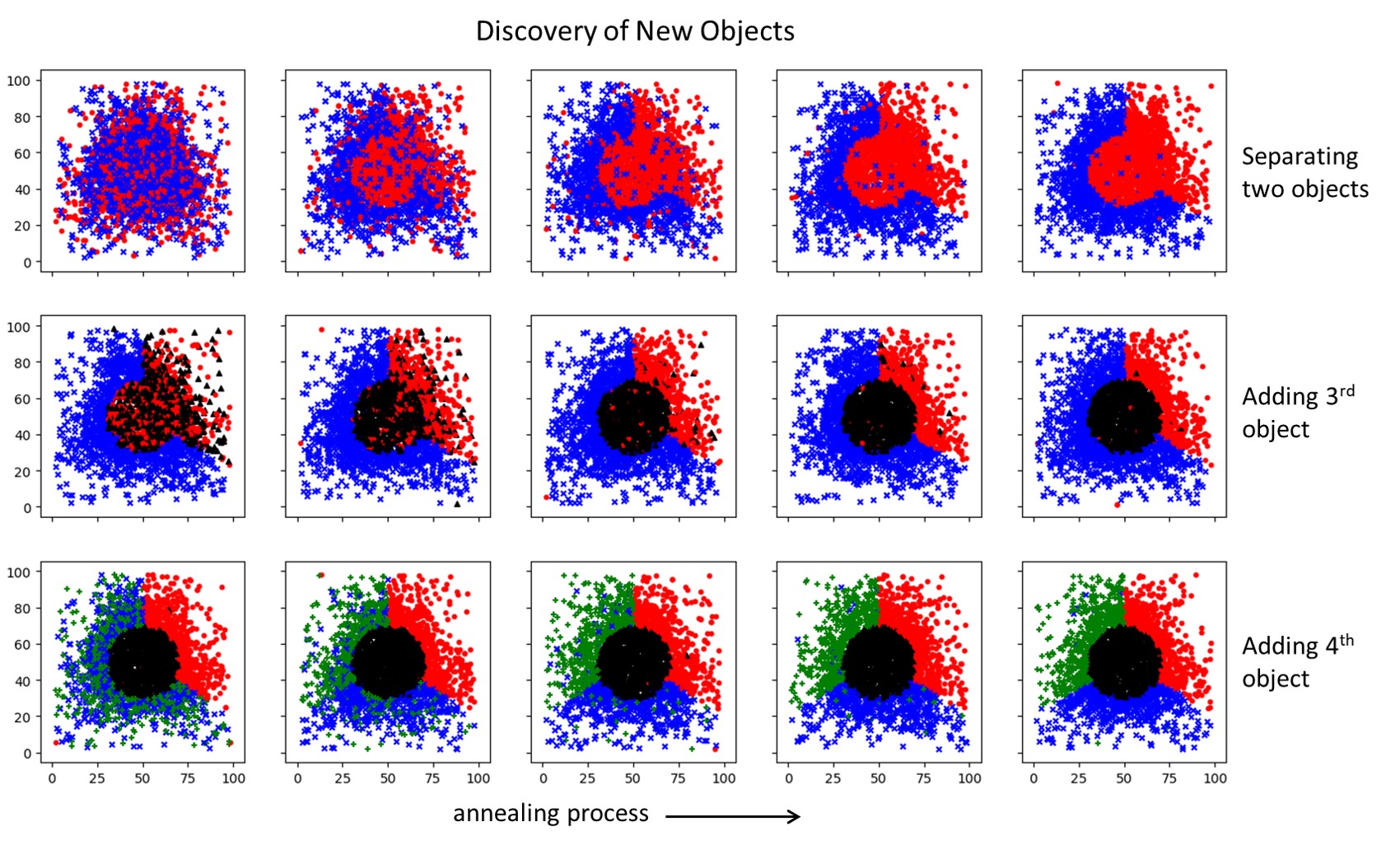}
    \caption{The machine learning algorithm is able to infer the existence of all four objects, along with their regions of operation.}
    \label{fig:img30}
\end{figure*}

The annealing process is visualized in Figure \ref{fig:img30}.  As shown in the figure, the program first discovers two suboptimal objects.  The program learns the correct spatial boundaries while determining the region of applicability for each object.  Notably, the regions of the datasets correlated to each object are connected spatially, but not temporally.  The ball's trajectory enters and exits each region multiple times during the simulation.  The algorithm does not enforce set connectedness or convexity during the annealing process, as seen in the starting positions on the left side of Figure \ref{fig:img30}.  The datapoints belonging to each set are initially dispersed randomly throughout the map.  They are iteratively sorted into their correct positions until the algorithm's error reaches a local minimum.

After this minimum is reached, the program hypothesizes the existence of a third object in the region of maximum error, which is the region indicated by red circles in the top row of Figure \ref{fig:img30}.  Predictably, the region containing the center of the map contains the maximum error, as the ball's motion in the center is more complex than near the edges.  The program repeats the annealing process on the second rows of Figure \ref{fig:img30} and correctly identifies the objects corresponding to Regions 1 and 4 in Figure \ref{fig:img61}a.

In the final iteration, shown on the third row of Figure \ref{fig:img30}, the program correctly identifies the objects corresponding to Regions 2 and 3.  The error in this state is the global minimum error of the system.  Adding a fifth object will not improve the model, nor will an attempt to unify existing objects as described on line 6 of Algorithm \ref{alg:structure_learning}.

We can also check the efficiency of the solution by observing the activation of the individual objects.  If we assume the ball has a fixed supply of energy to be distributed to the various objects, an accurate solution will show only one object active at each moment.  Figure \ref{fig:img56} shows the allocation of energy during the simulation.  As expected, the energy used by each object is either 100\% or 0\%, indicating that only one object is active at a time. The force vector in each region is shown in Figure \ref{fig:img61}d.
\section{Conclusion}

This work has described an object-relations technique as a method to understand and model general types of complex systems.  A mathematical description of the amount of information contained in complex systems is provided in the form of the Complex Information Entropy (CIE) equation.  Additionally, a Structure-Learning Algorithm is defined in order to infer the composition of complex systems, and a Zoom-in-Zoom-out (ZIZO) algorithm is provided in order to discover the connections among objects across a range of length- and timescales.  There are potentially far-reaching implications for these methods:
\begin{enumerate}
    \item In the field of ecology, researchers have recognized the species entropy of an ecosystem is correlated to its health, but have also recognized the imperfections in this method and have been searching for more accurate methods to measure environmental complexity.
    \item Comparable problems and applications exist in sociology, in regard to modelling the health of individuals in a community.
    \item Economists observe marketplaces are healthy when they contain a diverse array of small businesses.  In contrast to a capitalist economy, a tropenomy incentivized to optimize CIE could optimize the well-being of its participants without requiring constant growth.
    \item The continuous attractor networks used in neuroscience are an attempt to discover Pareto-optimal mathematical representations, in the form of compact topological spaces, to understand the coordinated activity of individual neurons relative to higher-order neural functions.
    \item In psychology, researchers are lacking in scientific methods to describe cognitive processes below surface-level observation.  There have been attempts in recent years to explain cognitive processes using theories based on entropy and thermodynamics, such as the Bayesian brain theory and the free energy principle, but these models assume the internal unity of the systems under investigation and do not account for modes of inner conflict, faulty reasoning, or emergence.
    \item In the field of biology, the CIE equation resolves the dark room paradox, which arises if one assumes an organism's nervous system is incentivized to minimize the entropy its predictive model of the world.  The organism could bring its predictive error to zero by sitting in a dark room and not observing anything.  The paradox is removed if one considers an organism's highest goal is to maximize CIE rather than to minimize prediction error, as this is equivalent to the organism developing the richest and most complete understanding possible of its environment, while maximizing $C$ includes minimizing object entropy as a subcomponent.
\end{enumerate}

Other potentially groundbreaking applications exist in the fields of engineering design, the physical sciences, and medicine.  As mentioned in Section \ref{section:ml}, if reinforcement learning is applied to a ZIZO-type algorithm, it creates the ability to design systems across multiple length- and timescales.  This creates the potential to generate complex engineering systems with an effectiveness beyond the capability of human experts.  Other possible applications exist in chemistry and materials science.  The ability to correlate molecular structure to macroscale chemical and material properties, and to design chemicals and materials with desired properties, is a major unsolved problem.

Similarly, the possibility exists to repair cells and tissues by using RL to discover cellular mechanisms of action, which are often unknown to medical researchers.  The development of treatments for medical conditions is difficult and time-consuming due to complex interactions present in the body.  The process often involves years of laboratory research and clinical testing.  In the long-term, the possibility exists to destroy individual cancer cells, rebuild cells in the nervous system, and make other highly specific and individualized cellular interventions.  Combining RL and active learning with the CIE equation creates a method to generate solutions to complex problems that hasn't been available previously.


\section*{Conflicts of Interest}

The author has no conflicts to disclose.

\section*{Data Availability Statement}

Data subject to third party restrictions.

\section*{References}

\bibliography{biblio}

\end{document}